\documentclass[lettersize,journal]{IEEEtran}
\usepackage{amsmath,amsfonts}
\usepackage{algorithmic}
\usepackage{algorithm}
\usepackage{array}
\usepackage[caption=false,font=normalsize,labelfont=sf,textfont=sf]{subfig}
\usepackage{textcomp}
\usepackage{stfloats}
\usepackage{url}
\usepackage{verbatim}
\usepackage{graphicx}
\usepackage{cite}
\begin{document}

\title{Predictive Traffic Rule Compliance using Reinforcement Learning}

\author{Yanliang Huang, Sebastian Mair, Zhuoqqi Zeng, Matthias Althoff
\thanks{This is a preprint of a paper intended for submission to IEEE ITSC 2025.}
}



\maketitle

\begin{abstract}
Autonomous vehicle path planning has reached a stage where safety and regulatory compliance are crucial. This paper presents an approach that integrates a motion planner with a deep reinforcement learning model to predict potential traffic rule violations. Our main innovation is replacing the standard actor network in an actor-critic method with a motion planning module, which ensures both stable and interpretable trajectory generation. In this setup, we use traffic rule robustness as the reward to train a reinforcement learning agent's critic, and the output of the critic is directly used as the cost function of the motion planner, which guides the choices of the trajectory. We incorporate some key interstate rules from the German Road Traffic Regulation into a rule book and use a graph-based state representation to handle complex traffic information. Experiments on an open German highway dataset show that the model can predict and prevent traffic rule violations beyond the planning horizon, increasing safety and rule compliance in challenging traffic scenarios.
\end{abstract}


\section{Introduction}
\IEEEPARstart{T}{he} field of autonomous driving has advanced substantially over the past five years. Although perception and prediction modules have become more reliable, planning systems still face challenges, particularly regarding safety assurance and operational robustness. Furthermore, traffic rule compliance remains a fundamental prerequisite for autonomous vehicles, both to protect road users and to satisfy legal certification standards.

Recent research has effectively applied temporal logic to formalize traffic rules, enabling automated online monitoring systems~\cite{Maierhofer2020,Maierhofer2022,Maler2004} to continuously monitor the compliance of traffic rules. These approaches use the concept of rule robustness—a quantitative metric indicating how thoroughly specific traffic rules are satisfied or violated. Integrating such monitors into autonomous driving frameworks can help prevent rule violations within the planning horizon.

However, a significant challenge emerges from the limited prediction horizons of autonomous driving modules, particularly in complex traffic scenarios. Current planning strategies, which respond solely to immediate violations, are unable to anticipate potential infringements beyond their present scope.

Building on steady progress in machine learning, there has been remarkable success in various autonomous driving applications, particularly in the interpretation of intricate traffic patterns and prediction of vehicle behaviors~\cite{Jordan2015}. Among various machine learning methods, reinforcement learning (RL) is especially suited for autonomous driving tasks due to its ability to adapt to complex environments, continuously learn and improve, and integrate rule information by setting proper rewards.

Building on these foundations, we propose to use an actor-critic (AC) framework that incorporates a hierarchically structured rule book comprising German interstate traffic rules. To facilitate predictive traffic rule compliance, we use the temporal logic robustness of these rules as reward signals, enabling the critic network to learn state-value functions that indicate potential rule violations. The learned value function then guides the actor’s exploration. 

Despite the benefits of deep reinforcement learning (DRL), agents often struggle with convergence in complex motion planning tasks involving multiple constraints. Although certain methods aim to enhance RL training efficiency~\cite{Peiss2023}, the outputs of actor networks can remain unstable, which requires extensive training for satisfactory performance.

To overcome these limitations, we propose a hybrid approach: we incorporate a cost-based motion planner as the actor component. In this thesis, we demonstrate our idea by adopting the Commonroad Reactive Planner~\cite{Wursching2024}, which has demonstrated both efficiency and reliability in generating robust, explainable trajectories. This planner leverages a lattice-based methodology~\cite{Werling2010,Werling2012} to produce candidate trajectories and identifies the optimal path via a cost function.

During inference, the motion planner reuses state-value estimates and incorporates additional cost factors to balance comfort, safety, and goal reachability, thereby enabling flexible configuration.

\begin{figure*}[htbp]
    \centering
    \includegraphics[width=0.8\textwidth]{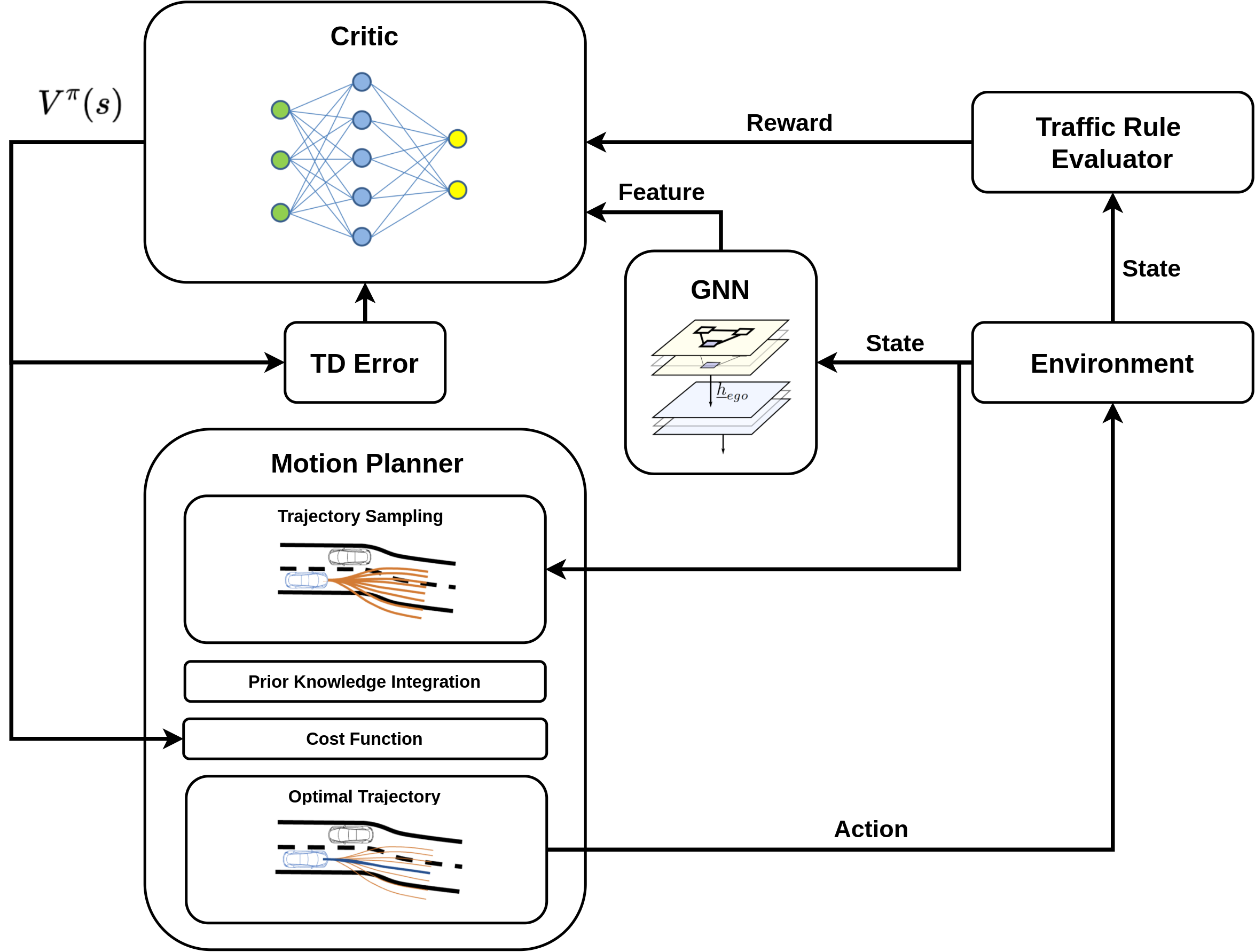}
    \caption{Overview of our proposed framework for traffic rule-aware motion planning.}
    \label{fig:overview}
\end{figure*}

Recognizing the importance of high-quality representations of environmental states in RL, we employ a Graph Neural Network (GNN) architecture~\cite{Meyer2023} as a feature extractor. This framework captures arbitrary road layouts and varying numbers of surrounding vehicles, while allowing flexible choices of input features. As a result, our agent benefits from a comprehensive state representation. An overview of the proposed approach is presented in Figure~\ref{fig:overview}.

In this paper, we present a new approach for predicting future traffic rule violations by integrating RL with motion planning for exploration. Our work makes several key contributions to the field of autonomous driving:

\begin{enumerate}
    \item We select three German interstate traffic rules that emphasize the importance of predictive traffic rule compliance.
    \item We pioneer the integration of AC with planning methods, creating a framework that combines the strengths of both.
    \item We use a graph-based model to efficiently extract and process complex traffic data, allowing robust feature representation.
    \item We validate our approach through extensive experiments on real-world traffic data, demonstrating its practical applicability and effectiveness.
\end{enumerate}

To the best of our knowledge, this is the first successful integration of AC algorithms with cost-based motion planning in autonomous driving. Our work opens new avenues for uniting learning-based prediction with explainable planning methods, particularly for forecasting and preventing future traffic rule violations.

\section{Background}
\subsection{Reinforcement Learning}

RL is a powerful machine learning paradigm that enables an agent to learn optimal decision-making strategies by interacting with an environment to maximize a cumulative reward signal. Unlike supervised learning, which depends on large, prelabeled datasets, RL adopts a trial-and-error approach, allowing the agent to mirror natural learning processes through direct environmental feedback. This framework is particularly well suited for solving Markov Decision Processes (MDPs), which formalize the environment in terms of states $s \in \mathcal{S}$, actions $a \in \mathcal{A}$, transition probabilities $P(s'|s,a)$, and rewards $r(s,a,s')$. The agent's goal is to derive a policy - a mapping from states to actions $\pi: \mathcal{S} \rightarrow \mathcal{A}$ - that optimizes the expected long-term reward~\cite{Sutton2018}.

To evaluate the quality of states and pairs of state actions, RL algorithms define value functions. The state value function $V^{\pi}(s)$ represents the expected return of the state $s$ following policy $\pi$.

One way to solve MDPs with RL is the AC method. It consists of two complementary components: The actor develops a policy $\pi(s)$ that generates actions, typically using a stochastic policy with a distributional output layer. The critic evaluates the quality of states through a value function $V(s)$, where $s$ denotes the system state~\cite{Sutton2018}.

In a standard AC implementation, the actor approximates the policy $\pi_\phi(a|s)$ and the critic approximates the state-value function $V_\xi(s)$, both using multilayer perceptrons (MLPs). These components maintain separate training processes but are updated simultaneously. The critic typically learns state values through methods like temporal difference (TD) learning or Monte Carlo estimation, providing feedback to refine the actor's policy.

Despite its promise, applying DRL to autonomous driving introduces challenges. The high-dimensional state spaces, multi-agent interactions, and the need to embed traffic rules and safety constraints into the learning process require tailored algorithmic advancements.

\subsection{Graph Representation and Graph Neural Network}

Graphs offer a powerful framework for modeling complex systems by representing entities as nodes and their relationships as edges. This structure excels at capturing topological relationships and dynamic interactions, making it particularly valuable in traffic scenarios where vehicles, pedestrians, and infrastructure elements interact. The graph is formally defined as $G = \{N, E\}$, where $N = \{n_i, i \in \{1, 2, ..., n\}\}$ represents the set of node attributes and $E = \{e_{ij}, i, j \in \{1, 2, ..., n\}\}$ denotes the set of edge attributes; $n$ represents the total number of vehicles in the constructed graph~\cite{Gkarmpounis2024}.

GNNs offer a framework for modeling the intricate relationships present in traffic environments, particularly the complex structures of the road network and dynamic vehicle interactions. Their effectiveness comes from the introduction of a relational bias to the learning problem through explicit modeling of connections between objects in the graph structure~\cite{Gkarmpounis2024}. 

\subsection{Traffic Rules Formalization}

Formalizing traffic rules into quantifiable metrics poses a significant hurdle for RL agents. Simple rules, such as speed limits, can be encoded directly through violation penalties, for example, a negative reward for exceeding a threshold. However, more sophisticated rules, such as \textit{"the ego vehicle must maintain sufficient speed to avoid impeding traffic flow,"} require comprehensive contextual information and cannot be adequately captured by binary compliance. Real-world driving often involves continuous degrees of adherence to the rules, and Boolean representations do not account for reasonable deviations, such as temporarily exceeding a speed limit to avoid an obstacle.

To overcome these limitations, formal temporal logic frameworks have emerged as powerful tools for specifying and evaluating traffic rules. Linear Temporal Logic (LTL) has been used to formulate traffic rules in an automatically evaluable syntax~\cite{Rong2020, Gao2019}. However, its discrete-time framework limits its suitability for online evaluation in RL. Signal Temporal Logic (STL), as an extension of Metric Temporal Logic (MTL)~\cite{Koymans1990}, addresses these shortcomings by specifying properties of continuous-time signals~\cite{Maler2004}. STL introduces quantitative semantics through the degree of robustness, which measures the degree of satisfaction or violation of the rule~\cite{Gressenbuch2021}. This makes STL ideal for autonomous driving.

\section{Methdology}

\subsection{Selection and Formulation of Traffic Rules}

This study centers on driving scenarios along the German interstate system, extending the foundational research presented in ~\cite{Maierhofer2020}. In that work, seven interstate traffic rules were formalized using STL, drawing from the German Road Traffic Regulation (StVO), the Vienna Convention on Road Traffic (VCoRT), and legal expertise. These rules address critical aspects of driving, including safety, speed limits, traffic flow, and local customs. However, not all of these formalized rules pose significant challenges or are equally relevant to our proposed methodology.

For instance, Rule $R\_G3$, which governs speed limits, can be incorporated into a motion planner through straightforward techniques. This can be achieved either by integrating a simple cost function that penalizes trajectories exceeding the speed limit or by filtering out non-compliant routes entirely. Similarly, Rule $R\_G2$ proves relatively trivial to enforce within our framework. Compliance can be ensured by evaluating the robustness of this rule across sampled trajectories within the planning horizon and selecting the optimal trajectory accordingly. Given the ease of adherence, violations of $R\_G2$ are infrequent and can be readily avoided using conventional planning methods.

Consequently, our approach emphasizes scenarios where conventional non-learning strategies are inadequate, highlighting the advantages of our data-driven, neural network-based methodology. For example, a vehicle with a 2 seconds prediction and planning horizon cannot effectively handle a rule violation at 2.1 seconds, since its motion planning and control systems typically cannot adjust in time, inevitably leading to a rule violation. To this end, we have carefully selected two representative rules from ~\cite{Maierhofer2020} that pose more complex challenges, requiring a sophisticated integration of traffic regulations into the autonomous vehicle decision-making process. Furthermore, we introduce a newly formalized rule to underscore the critical role of our method in addressing intricate traffic scenarios. This selection and formulation highlight the shortcomings of traditional approaches and illustrate the potential of our advanced techniques to improve autonomous driving performance on German interstates.

The rules are as follows:
\begin{itemize}
    \item \textbf{R\_G1 - Safe distance to preceding vehicle:} The ego vehicle following vehicles within the same lane must maintain a safe distance to ensure collision freedom, even if one or several vehicles suddenly stop. If another vehicle causes a safe distance violation due to a cut-in maneuver, the ego vehicle must recover the safe distance within a predefined time $t_c$ after the start of the cut-in ~\cite{Maierhofer2020}.
    
    \item \textbf{R\_I2 - Driving faster than left traffic:} The ego vehicle is not allowed to drive faster than any vehicle in the lanes to the left of it. Exception is the vehicle in the left lane is part of a queue of vehicles, slow-moving traffic, or congestion and the ego vehicle drives with only slightly higher speed ~\cite{Maierhofer2020}.
    
    \item \textbf{R\_I6 - No overtaking sign:} When the ego vehicle encounters a no overtaking sign, it must remain in the rightmost lane.
\end{itemize}

The detailed formulas for these rules, formalized using STL, are presented in Table~\ref{tab:rules}.

\begin{table*}
\caption{Overview of formalized traffic rules.}
\label{tab:rules}
\small 
\begin{tabular}{p{0.1\textwidth}p{0.3\textwidth}p{0.5\textwidth}}
\hline
\textbf{Rule} & \textbf{Law reference} & \textbf{STL formula} \\
\hline
R\_G1 & §4(1) StVO; §13(5) VCoRT & $G(\text{in\_same\_lane}(x_{\text{ego}}, x_0) \land \text{in\_front\_of}(x_{\text{ego}}, x_0)$ \\
& & $\land \neg \text{O}_{[t, t_c]} (\text{cut\_in}(x_0, x_{\text{ego}})$ \\
& & $\land \text{P}(\neg \text{cut\_in}(x_0, x_{\text{ego}})))$ \\
& & $\Rightarrow \text{keeps\_safe\_distance\_prec}(x_{\text{ego}}, x_0))$ \\
\hline
R\_I2 & §7(2), §7(2a), §7a StVO; & $G(\forall x_0 : \text{left\_of\_i}(x_0, x_{\text{ego}})$ \\
& ~\cite{Burmann2018} StVO §5 Rn. 58, & $\land \text{drives\_faster\_i}(x_{\text{ego}}, x_0)$ \\
& ~\cite{Burmann2018} StVO §18 Rn. 10-11 & $\Rightarrow (\text{in\_congestion}(x_0)$ \\
& & $\lor \text{in\_slow\_moving\_traffic}(x_0)$ \\
& & $\lor\text{in\_queue\_of\_vehicles}(x_0)))$ \\
\hline
R\_I6 & §2(1), §5(1)(3) StVO; & $G(\text{no\_overtaking\_sign}(x_{\text{ego}})$ \\
& traffic sign 276 & $\rightarrow \text{in\_rightmost\_lane}(x_{\text{ego}}))$ \\
\hline
\end{tabular}
\end{table*}

We select rule $R\_G1$, the safe distance rule, as the initial test of the validity of our approach. For a motion planner, compliance with this rule is generally straightforward, with robustness cost in trajectory selection sufficing in most cases. However, when a preceding vehicle brakes abruptly due to congested traffic, the planner may fail to respond quickly enough, resulting in a violation. In contrast, our RL-based method leverages a comprehensive view of the scenario within the sensor range to predict other agents' braking, enabling proactive decision-making to maintain a safe distance.

Furthermore, we simplified rule $R\_I2$, omitting certain exception conditions present in ~\cite{Maierhofer2020}, as our dataset lacks broad lane markings and ramps, but the simplified rule retains its relevance. Our RL method may outperform approaches that rely solely on the cost of robustness, particularly when a slow-moving vehicle in the left lane exhibits a significant velocity difference relative to the ego vehicle. In such cases, a predictive approach can exploit additional information to initiate maneuvers earlier.

\begin{figure}[htbp]
    \centering
    \includegraphics[width=\columnwidth]{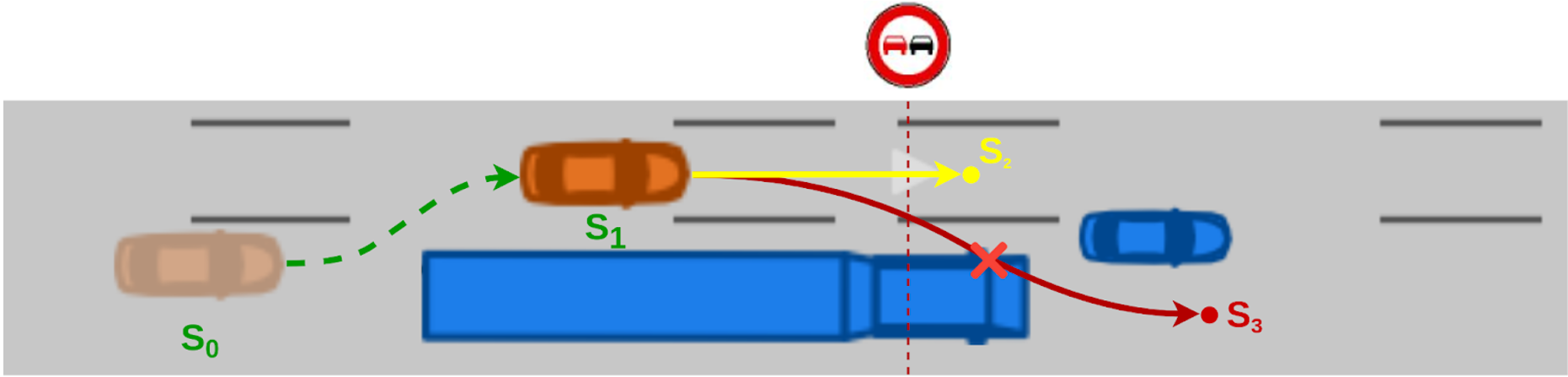}
    \caption{A scenario where predictive methods are needed}
    \label{figure:R_I6}
\end{figure}

Beyond the rules adopted directly from ~\cite{Maierhofer2020}, we formalize a new rule, $R\_I6$, based on German traffic regulations. According to §2(1) of the StVO, on multi-lane roadways in one direction, vehicles may deviate from the obligation to drive as far right as possible. Section §5(1) mandates overtaking on the left lane, while §5(3) prohibits overtaking when indicated by traffic signs (e.g., Sign 276 or 277). Thus, for maximum compliance, a vehicle should remain in the rightmost lane when passing a \textit{no overtaking} sign. Although simple, this rule presents challenges in specific scenarios. 

Consider, for example, a situation where an ego vehicle is overtaking a long truck on a two-lane road. A planning method based solely on robustness cost would maintain compliance with $R\_I6$ until the vehicle is near the traffic sign. However, as the vehicle approaches the sign, there may be insufficient time and space to complete the overtaking maneuver or safely return to the right lane. As illustrated in Figure~\ref{figure:R_I6}, the green arrow represents a rule-compliant trajectory, the yellow arrow indicates a rule-violated trajectory, and the red arrow denotes an infeasible trajectory. Within its immediate planning horizon, the ego vehicle can maintain compliance with traffic rules. However, it is unable to reposition itself into the right lane before encountering the \textit{no overtaking} sign. This constraint inevitably leads to a rule violation, exposing a key limitation of traditional planning approaches: breaches may become unavoidable when the planning horizon fails to anticipate downstream constraints.

In contrast, a predictive approach offers a robust solution. By detecting the \textit{no overtaking} sign within the vehicle sensor range, the system can preemptively maneuver to the right lane well before reaching the prohibited zone. This strategy prevents further overtaking attempts as the vehicle approaches the sign, ensuring full compliance with $R\_I6$.

To address potential rule conflicts, we implemented a hierarchical rule book structure~\cite{Censi2019} that reflects the natural priority ordering of traffic rules. This structure acknowledges that lower-priority rules may need to be temporarily violated to satisfy higher-priority rules in challenging situations. $R\_G1$ receives highest priority due to its direct impact on collision avoidance and safety-critical interactions. $R\_I6$ follows as the second-highest priority due to its role in maintaining traffic order and preventing dangerous overtaking behaviors. $R\_I2$ receives the lowest priority as it primarily affects traffic efficiency rather than immediate safety concerns. Following this prioritization, we establish the rule book hierarchy:

$$ R\_G1 > R\_I6 > R\_I2 $$

\subsection{Graph Representation and GNN}

We construct a graph representation of the traffic scenario. A graph consists of nodes and edges. In the context of traffic, the nodes could represent vehicles, lanes, or other participants such as bicycles and pedestrians. However, given the location on the highway, we exclude non-vehicle participants. Furthermore, the relatively simple layout of the lane of the roads, mainly consisting of roads and ramps, makes the use of lanelets unnecessary. Representing lanelets as nodes would introduce multiple node types, resulting in a heterogeneous graph, which is more complex to process with GNNs. Consequently, we simplify the representation by using only vehicle nodes connected by vehicle-to-vehicle (V2V) edges to form the graph. 

In \cite{Hart2020}, the $V2V$ graph is constructed using basic kinematic vehicle features and relative position edge features, as detailed in Table~\ref{tab:graph-features}. However, these features alone are insufficient for the model to develop a comprehensive understanding required to predict compliance with traffic rules. To address this limitation, we enhance the graph representation by incorporating additional features designed to help the model recognize specific patterns associated with rule compliance.

Drawing inspiration from ~\cite{Peiss2023}, which employs rule robustness as node features and predicates such as \textit{CutIn} and \textit{InFrontOf} as edge features, we initially considered a similar approach. However, the computational overhead of repeatedly extracting features and calculating robustness proved prohibitive for our method, which involves numerous iterative operations. Instead, we adopt simpler, yet targeted, features tailored to our specific traffic rules. The enhanced graph representation, including these additional features, is also presented in Table~\ref{tab:graph-features}.

\begin{table}
\centering
\caption{Graph Feature definitions}
\label{tab:graph-features}
\begin{tabular}{ccc}
\hline
\textbf{Feature} & \textbf{Ego Feature name} & \textbf{Normalization function} \\
\hline
\multicolumn{3}{c}{\textit{Existing vehicle node features}} \\
\hline
$f_{p,x}$ & PosEgoFrameX & $f_{p,x} / 50$ \\
$f_{p,y}$ & PosEgoFrameY & $f_{p,y} / 50$ \\
$f_v$ & Velocity & $(f_v - 15) / 20$ \\
\hline
\multicolumn{3}{c}{\textit{Existing ego vehicle features}} \\
\hline
$f_a$ & Acceleration & $f_a / 20$ \\
$f_v$ & Velocity & $(f_v - 15) / 20$ \\
$f_y$ & YawRate & $\min(\max(f_y,-1),1)$ \\
\hline
\multicolumn{3}{c}{\textit{Existing Vehicle-to-vehicle edge features}} \\
\hline
$f_{rp,x}$ & RelativePosEgoX & $f_{rp,x} / 50$ \\
$f_{rp,y}$ & RelativePosEgoY & $f_{rp,y} / 50$ \\
\hline
\multicolumn{3}{c}{\textit{Added vehicle node features}} \\
\hline
$f_{l,L}$ & Lane & $f_{l,L}$ \\
\hline
\multicolumn{3}{c}{\textit{Added ego vehicle features}} \\
\hline
$f_{lb,L}$ & DistLeftBound & $f_{lb,L} / 2$ \\
$f_{rb,L}$ & DistRightBound & $f_{rb,L} / 2$ \\
$f_{lrb,L}$ & DistLeftRoadBound & $(f_{lrb,L} + f_{lb,L}) / 12$ \\
$f_{rrb,L}$ & DistRightRoadBound & $(f_{rrb,L} + f_{rb,L}) / 12$ \\
$f_{h,L}$ & HeadingError & $\min(\max(f_{h,L},-\pi/4),\pi/4)$ \\
$f_{la,L}$ & GoalDistLateral & $\log(|f_{la,L}| + 1) f_{la,L} / |f_{la,L}|$ \\
$f_{lo,L}$ & GoalDistLongitudinal & $\log(|f_{lo,L}| + 1) f_{lo,L} / |f_{lo,L}|$ \\
$f_{l,L}$ & Lane & $f_{l,L}$ \\
$f_{nots,L}$ & NonOvertakingTSRelative & ($f_{nots,L}-50) / 50.0$ \\
\hline
\multicolumn{3}{c}{\textit{Added Vehicle-to-vehicle edge features}} \\
\hline
$f_{rv,x}$ & RelativeVelocityEgoX & $f_{rv,x} / 20.0$ \\
$f_{rv,y}$ & RelativeVelocityEgoY & $f_{rv,y} / 20.0$ \\
$f_{lo,LT}$ & LeftOf & $f_{lo,LT}$ \\
$f_{sl,LT}$ & SameLane & $f_{sl,LT}$ \\
\hline
\end{tabular}
\end{table}

Intuitively, the ego vehicle has access to more detailed information than other agents. Therefore, in addition to standard node and edge features, we incorporate extensive ego-specific features, which are later combined with the embedded features from the GNN. Key additions include features such as \textit{DistLeftBound}, \textit{DistRightBound}, \textit{DistLeftRoadBound}, and \textit{DistRightRoadBound}, which are added to both vehicle and ego features to enable vehicles to localize themselves within the lanelet network. Since lanelets are not represented as nodes, these features are essential for recognizing lane-related rules, all of which are relevant to our selected rules. Additionally, a \textit{Lane Feature} explicitly indicates the lane occupied by each vehicle, facilitating the assessment of rule compliance. For traffic rule $R\_{I6}$, we introduce the \textit{NonOvertakingTSRelative} feature, which provides the relative longitudinal distance to the next \textit{``no overtaking''} traffic sign, based on a sensor radius of 100 meters. This range ensures the vehicle has sufficient time to maneuver and avoid violations. The distance resets to the next relevant sign upon encountering a \textit{``no overtaking end''} sign. Furthermore, \textit{HeadingError} and goal-related features enhance the ego vehicle's situational awareness and support goal-oriented planning.

For $V2V$ edge features, we introduce the \textit{RelativeVelocity} feature, which is critical for ensuring the ego vehicle does not exceed the speed of the vehicle to its left, supporting compliance with rule $R\_{I2}$. The \textit{LeftOf} feature mimics the predicate calculation for \textit{``LeftOf,''} providing expressive information with low computational complexity. Additionally, the \textit{SameLane} feature aids in understanding rule $R\_G2$, which relates to interactions with vehicles in the same lane.

By integrating these features, we provide the model with rich, context-aware information to enable accurate predictions of traffic rule compliance.

\subsection{Modified Actor-Critic Algorithm}

Applying AC method to autonomous driving presents significant challenges. When a neural network serves as the actor, its black-box nature raises concerns about interpretability and controllability. Integrating prior knowledge, such as collision avoidance, becomes difficult, leading to unsafe or unstable vehicle behaviors that are impossible to debug or constrain. Furthermore, a randomly initialized neural network actor begins by making random decisions, which impedes effective exploration. This issue is particularly pronounced in interstate traffic scenarios, where the navigable space is narrow relative to the length of the route. Random actions often result in off-road terminations, preventing the agent from reaching its goal.

To address these limitations, we propose replacing the traditional actor network with a motion planner equipped with a cost function. This planner incorporates critical prior knowledge on trajectory feasibility and safety, including kinematic constraints. Ensures physically feasible trajectories by verifying the kinematic feasibility and collision avoidance before selecting the optimal trajectory based on a predefined cost function. Unlike a neural network, the planner's decision-making process is interpretable and adjustable. By integrating a series of modules, we can embed prior knowledge into the planner, enabling it to explore the environment more effectively across diverse scenarios.

The effectiveness of this replacement stems from the fact that the output of the value network approximates the state value, which represents the cumulative rewards following the policy. When using robustness of traffic rules as a reward, a higher state value indicates that the vehicle is more likely to comply with rules in that state, while a lower value suggests the opposite, which is essential to avoid long-term violations of traffic rules. Consequently, in the exploration phase of our approach, the motion planner selects trajectories using a state-value-guided mechanism.

This selection process identifies a "macro-action", a trajectory spanning multiple time steps that balances long-term value with prior knowledge. The motion planner constrains the action space to a finite set of trajectories satisfying kinematic and safety requirements. By incorporating $-V^*(s_{ij})$ into the cost function, an estimate of future returns is included for each state $s_{ij}$ along the trajectory $\tau_i$. The critic then updates its value network based on rollouts collected through the actor's exploration. This approach aligns with the AC framework's core principle: the actor selects actions according to the current policy, the critic evaluates the outcomes and estimates their quality, and the actor refines its policy using the critic's feedback.

Moreover, during the exploration phase, rather than employing a $\varepsilon$-greedy strategy, a greedy strategy is adopted with random start and goal settings and a planned route. Since the motion planner generates trajectories around the route, the created routes with random settings ensure comprehensive coverage of possible states.

Due to the simultaneous updates of both the actor and critic, and since the critic's state value updates depend on the next states produced by the current policy, this method achieves on-policy learning while maintaining high sample efficiency.

In contrast, our approach allows a trained critic to represent the state-value for one or multiple rewards, enabling the combination of different critic networks. This functions as a plug-and-play module within any cost-based planner, allowing a motion planner to leverage multiple critic networks as cost calculators simultaneously to pursue multiple planning goals. The only consideration is to assign appropriate weights to balance the trade-offs among them.

\section{Experiments}

This chapter details our experimental investigation, spanning the foundational setup, implementation details, and a comparative analysis of our modified RL approach against a baseline model. We leverage the highD dataset for evaluation. All experiments were conducted on a computing platform equipped with 12 CPUs, an Nvidia RTX 2060 GPU, and 30~GB of RAM. The training process was organized into three distinct phases, each targeting a specific traffic rule and comprising 20,000 steps.

\subsection{Dataset}

Our study employs the highD dataset~\cite{Krajewski2018}, a comprehensive collection of naturalistic vehicle trajectories captured via drone-based aerial observation on German highways. It encompasses 1,255 traffic scenarios, each lasting 40--50 seconds and recorded at 25~Hz (0.04~s intervals) over 1,000 timesteps, presented in Bird's Eye View (BEV) format. This mid-level representation provides detailed trajectory data, including vehicle specifications, dimensions, and maneuver patterns.

\subsection{Experimental Setup}

For the experiment, we selected 100 scenarios from the highD dataset. Given that no \textit{no overtaking} signs are present in the original data, we manually inserted one into each scenario, randomly positioned between 100m and 350m along the longitudinal axis, to simulate a variety of situations.

The hyperparameters for the environment, model, and training are presented in Table~\ref{tab:model-hyperparams}.

\begin{table}
\centering
\caption{Environment, model and training hyperparameters}
\label{tab:model-hyperparams}
\begin{tabular}{ll}
\hline
\textbf{Parameter} & \textbf{Value} \\
\hline
\multicolumn{2}{c}{\textit{Environment}} \\
\hline
Interval (s) & 0.1 \\
Ego vehicle start state $(l_s, v_s)$ & 150~m to 350~m from the goal, 15 m/s \\
Minimum Distance to Vehicles (m) & 3 \\
\hline
\multicolumn{2}{c}{\textit{Model}} \\
\hline
V2V graph construction & 3-nearest neighbor $\forall\|\Delta p_{vw}\|_2 < 50$ \\
GNN message passing layers $(K)$ & 3 \\
Aggregation, activation functions & $\max(\cdot)$, $\tanh(\cdot)$ \\
Feature dimensions $(|\mathbf{h}_v^{k>0}|, |\mathbf{z}_t|)$ & 80, 80 \\
GNN ego node-feature embedder & $MLP(|\mathbf{h}_{ego}| + |f_{ego,LT}|, |\mathbf{z}_t|)$ \\
AC networks $(\pi, V)$ & Each $MLP(256, 128, 64)$ \\
\hline
\multicolumn{2}{c}{\textit{Motion Planner}} \\
\hline
Replanning Frequency (Hz) & 2 \\
Planning Horizon (s) & 2 \\
Sampling levels & 2 \\
Mode & Velocity Keeping \\
\hline
\multicolumn{2}{c}{\textit{Learning}} \\
\hline
Rollout steps & 256 \\
Discount factor $(\gamma)$ & 0.99 \\
\hline
\multicolumn{2}{c}{\textit{Training}} \\
\hline
Optimizer & Adam \\
Learning rate & $5 \times 10^{-4}$ \\
Weight decay & $10^{-3}$ \\
Number of epochs & 8 \\
Batch size & 32 \\
\hline
\end{tabular}
\end{table}

Exploration is guided by the CommonRoad Reactive Planner~\cite{Wursching2024}, which operates with a replanning interval of 0.5~s and generates a set of trajectories spanning 20 timesteps (i.e., a 2~s horizon).

\subsection{Rewards Design}

We implement an independent training phase for each traffic rule, employing rule robustness measures as reward signals with the same features, agent, and model. This separation facilitates focused learning of individual rule characteristics, minimizing cross-interference and eliminating the need to define and tune weights before training, thereby streamlining the process.

Moreover, we identified a challenge due to the absence of a termination reward. Without a positive reward for reaching the destination, and with a sparse or zero reward function throughout an episode, the cumulative future reward diminishes as the ego vehicle approaches the goal. This reduction occurs because fewer remaining steps limit opportunities to accumulate rewards, resulting in a counterintuitive decrease in the state-value function $V(s)$ near the destination. To address this, we introduce a \textit{TrajectoryProgressionRewardComputer}, which provides positive rewards for advancing toward the goal, ensuring that the agent learns a policy that efficiently prioritizes its achievement. The reward encourages progression toward the goal by rewarding both the extent and the rate of advancement. As the vehicle progresses, the arc length increases, contributing positively to the reward through the term $(1 - \text{dynamic\_weight}) \times \text{arclength}$; even if progress slows near the goal, this term grows as the arc length approaches its maximum.

\begin{table}
\centering
\caption{Reward computer for R}
\label{tab:reward-computer}
\begin{tabular}{ll}
\hline
\textbf{Reward computer} & \textbf{Weight} \\
\hline
\multicolumn{2}{c}{\textit{Phase 1}} \\
\hline
I2RobustnessRewardComputer & 10 \\
TrajectoryProgressionRewardComputer & 8 \\
\hline
\multicolumn{2}{c}{\textit{Phase 2}} \\
\hline
I6RobustnessRewardComputer & 10 \\
TrajectoryProgressionRewardComputer & 8 \\
\hline
\multicolumn{2}{c}{\textit{Phase 3}} \\
\hline
G1RobustnessRewardComputer & 10 \\
TrajectoryProgressionRewardComputer & 8 \\
\hline
\end{tabular}
\end{table}

\subsection{Benchmarking}

To assess our approach, we established a non-learning baseline motion planner. This baseline substitutes learned value predictions with immediate rule evaluations via online robustness monitoring. Consequently, it lacks the predictive capacity to foresee and mitigate rule violations beyond the current window.

\subsection{Evaluation}

We use the explained variance metric and the episode reward mean metric to monitor and evaluate training.

\subsubsection{Explained Variance}

The explained variance measures how well the critic value function $V_\phi(s)$ predicts the actual returns observed during training. It is defined as:

\begin{equation}
\text{Explained Variance} = 1 - \frac{\text{Var}(G_t - V_\phi(s_t))}{\text{Var}(G_t)}
\end{equation}

where $G_t$ is the true return, $V_\phi(s_t)$ is the predicted value, and $\text{Var}$ denotes variance. A value close to 1 indicates that the critic accurately captures the variability in returns, while a value close to 0 or negative suggests poor prediction.

\begin{figure}[htbp]
    \centering
    \includegraphics[width=0.8\columnwidth]{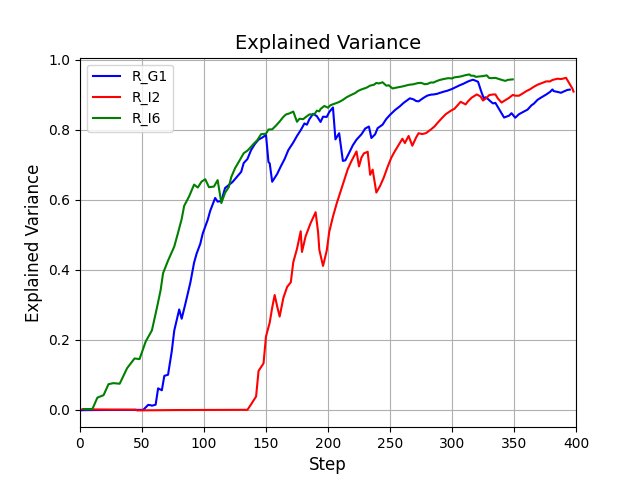}
    \caption{Explained Variance}
    \label{fig:explained_variance}
\end{figure}

In Figure~\ref{fig:explained_variance}, the explained variance graph illustrates distinct learning dynamics for the traffic rules over 400 training steps. Specifically, $R\_{G1}$ demonstrates a steady increase, rising from 0 to approximately 0.9. This consistent ascent is particularly pronounced up to step 200, after which it stabilizes with minor fluctuations, reflecting the model's progressively improving ability to predict returns. In contrast, $R\_{J2}$ exhibits greater volatility, characterized by notable peaks and troughs, while $R\_{J6}$ shows a slower, more gradual increase. The steady progress of $R\_{G1}$ indicates that the model effectively enhances its capacity to ensure long-term compliance with the safe distance rule, aligning with our evaluation objectives. Additionally, $R\_{I6}$ emerges as the easiest rule to learn, as it involves no vehicle-to-vehicle interactions, whereas $R\_{I2}$ proves the most challenging due to the extensive vehicle interactions it requires the model to capture.

\subsubsection{Episode Reward Mean}

The episode reward mean is the average total reward collected over an episode, a complete driving scenario from start to finish. For our setup, this likely reflects the cumulative robustness or compliance with traffic rules like $R\_{G1}$, averaged across multiple episodes:

\begin{equation}
\text{Episode Reward Mean} = \frac{1}{N} \sum_{e=1}^N \sum_{t=0}^{T_e} r_t^e
\end{equation}

where $N$ is the number of episodes, $T_e$ is the length of episode $e$, and $r_t^e$ is the reward in time step $t$ in episode $e$. This metric evaluates the policy's overall performance. A higher mean reward suggests that the ego vehicle is consistently achieving better outcomes.

\begin{figure}[htbp]
    \centering
    \includegraphics[width=0.8\columnwidth]{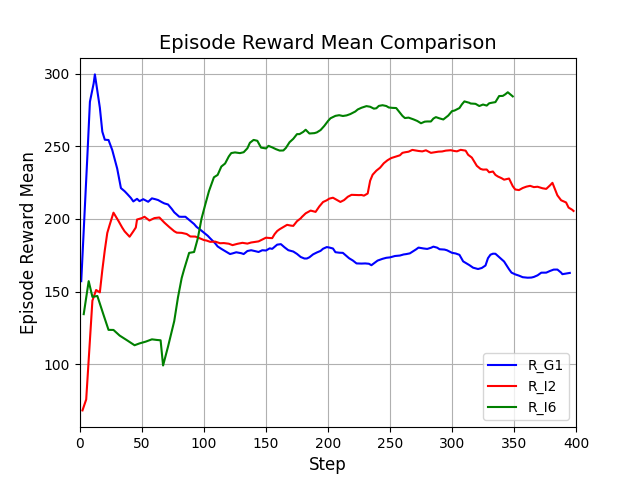}
    \caption{Episode Reward mean}
    \label{fig:erm}
\end{figure}

The \textit{Episode Reward Mean Comparison} graph evaluates three configurations over 400 steps, with rewards ranging from 0 to 300. The $R\_{G1}$ starts at around 300 but drops to 150 by step 50, stabilizing thereafter, possibly due to early overfitting. The $R\_{G2}$ begins at 50, peaks at 250 by step 300, and fluctuates between 150-250, indicating acceptable training process. The $R\_{I6}$ starts near 0 and rises steadily to 300 by step 400, surpassing the others with a robust upward trend. This suggests that $R\_{I6}$ achieves the highest long-term reward, reflecting better adaptability and learning efficiency.

\subsubsection{Performance of R\_I6 Predictive Model}

\begin{figure*}[htbp]
    \centering
    \includegraphics[width=0.8\textwidth]{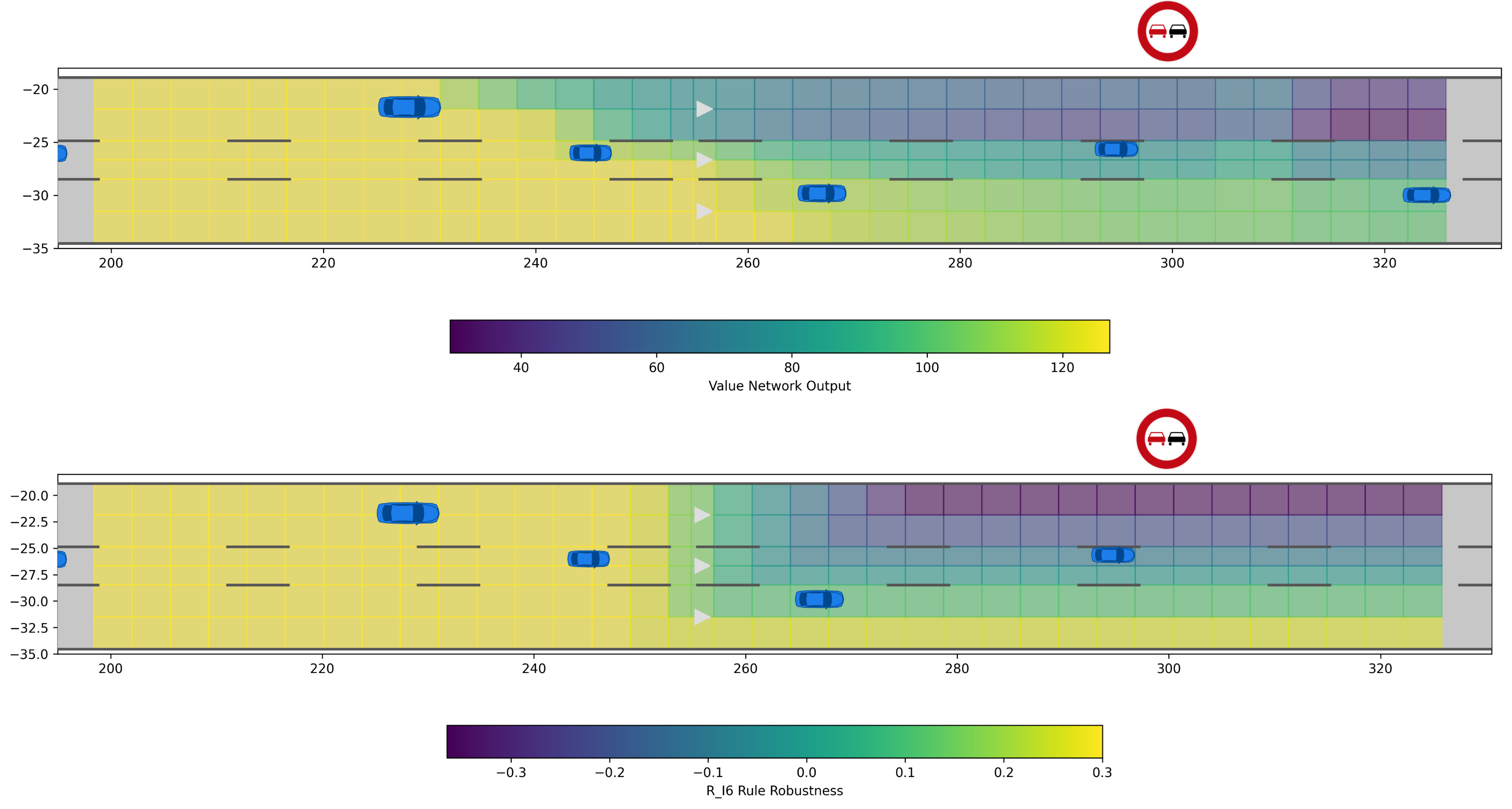}
    \caption{Heatmap comparison of state-value and rule robustness for $R\_I6$}
    \label{fig:R_I6_result_comparison}
\end{figure*}

Our primary focus is on the custom rule $R\_{I6}$, for which we compare the trained value network with the baseline model within the same scenario, featuring a \textit{no overtaking} sign positioned at 300m along the longitudinal axis. We generate heatmaps using a grid-segmented representation of the road, evaluating the approximated state-value and $R\_{I6}$ rule robustness for both models. As depicted in Figure~\ref{fig:R_I6_result_comparison}, a yellow grid indicates higher values, while a purple grid signifies lower values. The figure presents a comparative analysis of the trained value network and the baseline model, utilizing heatmaps overlaid on the road grid to assess the approximated state-value and $R\_{I6}$ rule robustness.

The bottom baseline model heatmap displays a sharp yellow-to-purple transition between 250m and 300m, attributable to the traffic sign detection range being set at 50~m. This transition forms a rectangular shape, as the model increasingly avoids leftward positions as it approaches the sign. This suggests that the baseline model relies heavily on immediate rule evaluation, with limited foresight into the \textit{no overtaking} constraint.

The trained value heatmap exhibits a smooth transition from yellow to purple, approximately aligning with the baseline model. However, the initial value change begins earlier, around 230m, compared to the 250m detection range; the range increased from 50m to around 70m, indicating an awareness of long-term rule violation risks. Notably, the heatmap adopts a trapezoidal shape, with the value in the left lane decreasing earlier than in the right lane. Leveraging the predictive model, the lateral position of the ego vehicle influences the timing of lane-changing maneuvers: a vehicle positioned farther to the left can initiate a rightward lane change earlier, whereas a vehicle already near the right lane may delay the maneuver without violating the rule. This behavior aligns with human intuition and enhances safety and efficiency, as it enables proactive adjustments tailored to the vehicle’s spatial context.

The alignment between the robustness and state-value heatmaps of both models reinforces that 
rule robustness directly shapes the value function. However, the trained model’s smoother transitions suggest superior integration of long-term planning.

\subsubsection{Performance of R\_G1 Predictive Model}

\begin{figure*}[htbp]
    \centering
    \includegraphics[width=0.8\textwidth]{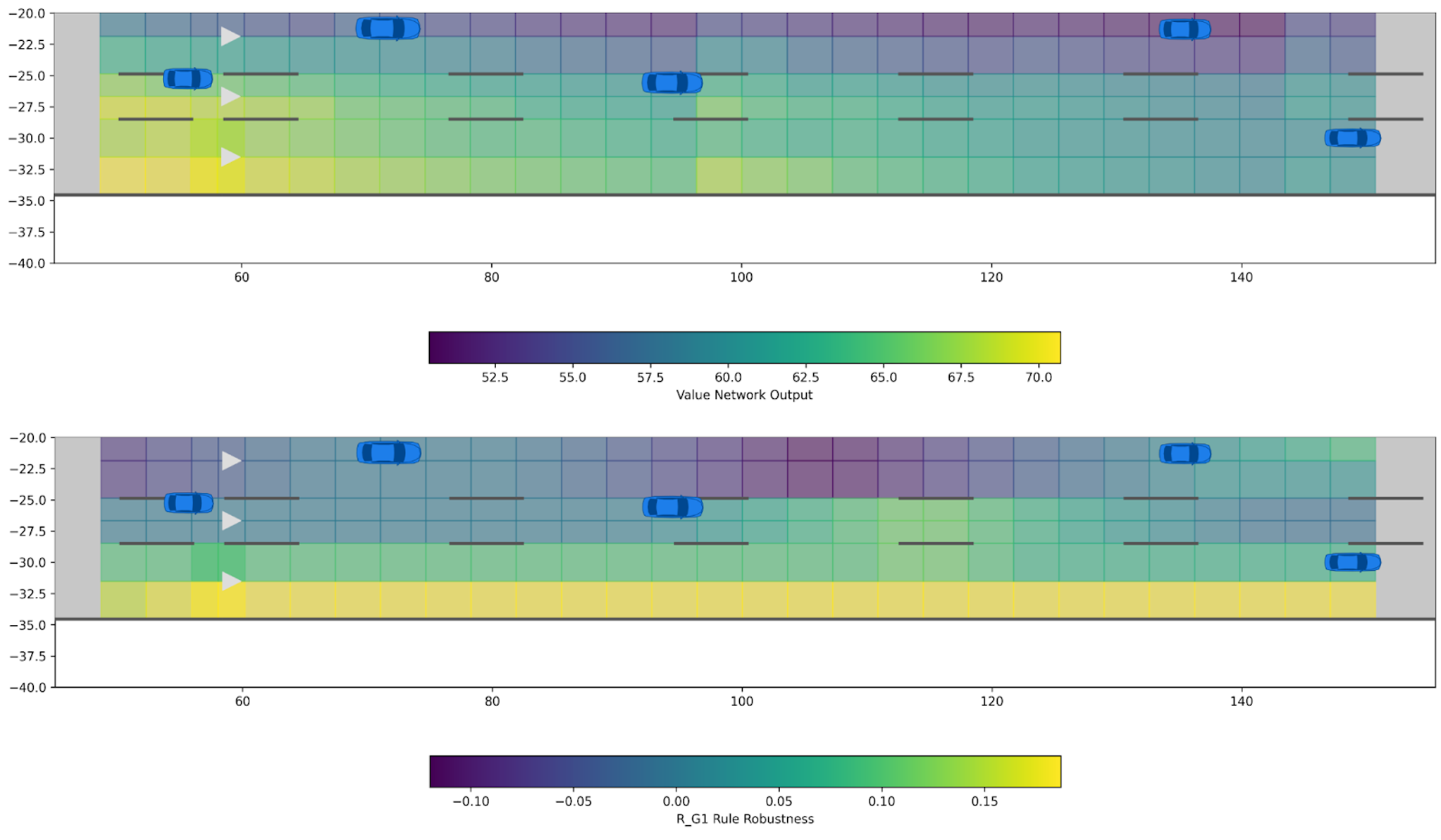}
    \caption{Heatmap comparison of state-value and rule robustness for $R\_G1$}
    \label{fig:R_G1_result_comparison}
\end{figure*}

The robustness of the value network output for rule $R\_{G1}$ is illustrated in Figure~\ref{fig:R_G1_result_comparison}, where the speed of the ego vehicle is set to 25\,m/s to simulate a vehicle approaching rapidly. In the bottom heatmap depicting the robustness of the rule, the positions directly behind each vehicle exhibit lower robustness, with the area behind the vehicle in the left lane showing the lowest values. Robustness remains higher in the right lane, particularly at greater lateral distances from vehicles, suggesting that maintaining an offset reduces collision risk.

The upper heat map, which represents the output of the value network, shows broader and larger unsafe regions behind vehicles compared to the robustness heat map. For example, for the vehicle at $(150 m, -31 m)$, the robustness heatmap indicates low values only in the grids directly behind it. In contrast, the output of the value network increases in adjacent lateral positions as the ego vehicle approaches, reflecting consideration of possible lane changes. This adjustment allows the model to account for dynamic scenarios, such as sudden maneuvers by other vehicles, and supports a more conservative policy to prevent $R\_G1$ violations. The value network's ability to incorporate longer-term predictions distinguishes it from the baseline model, enhancing its effectiveness in managing variable traffic conditions.

\subsubsection{Performance of R\_I2 Predictive Model}

\begin{figure*}[htbp]
    \centering
    \includegraphics[width=0.8\textwidth]{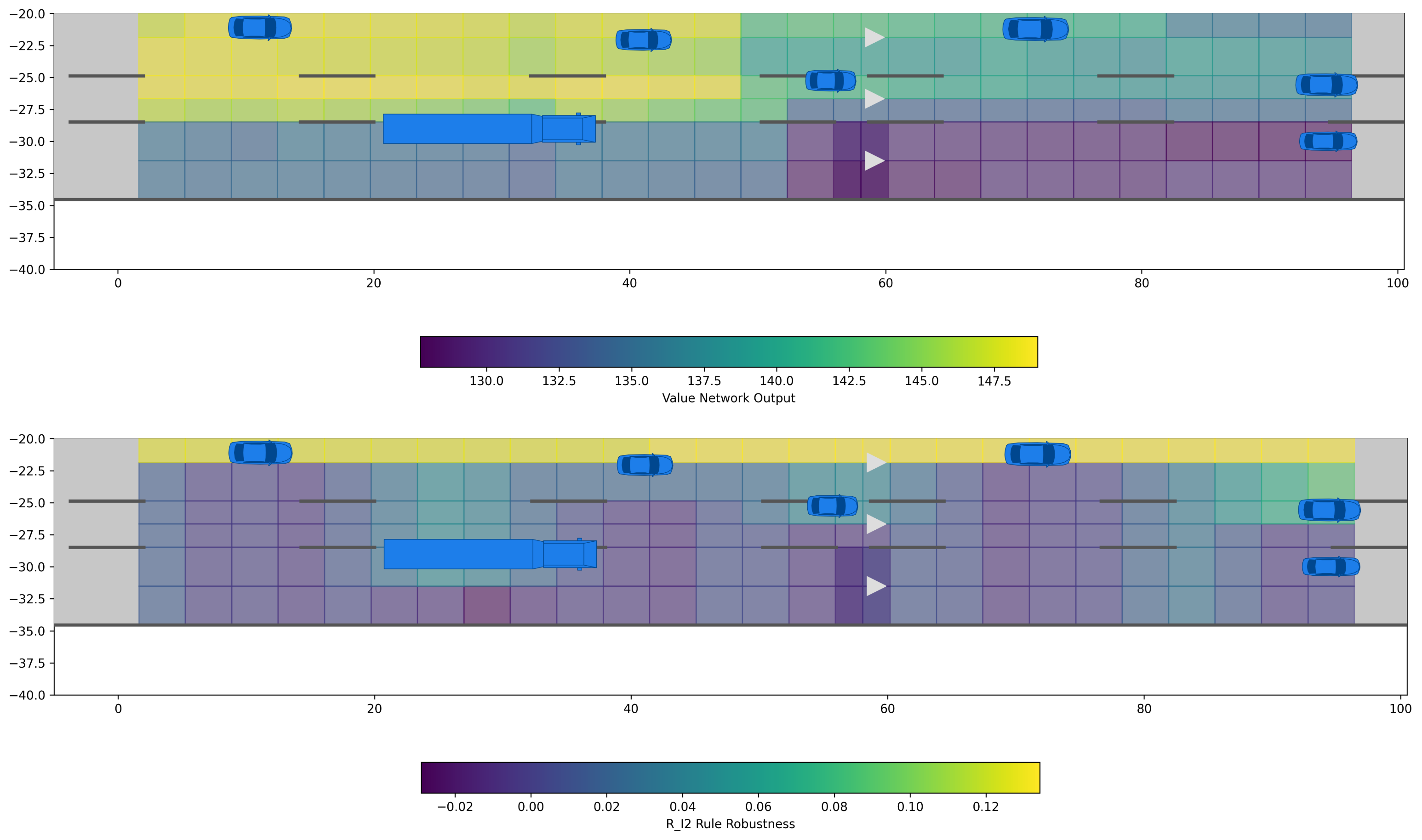}
    \caption{Heatmap comparison of state-value and rule robustness for $R\_I2$}
    \label{fig:R_I2_result_comparison}
\end{figure*}

In Figure~\ref{fig:R_I2_result_comparison}, the speed of the ego vehicle is set to 25\,m/s. The robustness heatmap indicates that the positive robustness is limited to the leftmost lane and specific grids in the middle lane where gaps between vehicles occur. This pattern highlights that 25\,m/s exceeds the speed of most surrounding vehicles, restricting rule-compliant positions to areas where the ego vehicle avoids direct adjacency to slower vehicles on its left. In the value heatmap, which represents the model's evaluation of long-term compliance and reward potential, the right lane consistently exhibits low values, aligning with the robustness findings. In contrast, the left lane shows higher values in the first half of the road compared to the second half. This variation is attributed to a slower moving vehicle at coordinates (75, -21), increasing the risk of violating the rule $R\_{I2}$ when the ego vehicle travels along it. Such insights could enable the vehicle to make more informed and sensitive decisions. However, the prediction of values shows significant volatility, consistent with the analysis of ``Explained Variance'', underscoring the learning challenges associated with $R\_{I2}$. This volatility suggests that additional training data and extended learning periods are essential to improve the stability and predictive accuracy of the model.

\subsection{Discussion}

This experiment demonstrates the model's capability to predict long-term violations of traffic rules within the highD dataset scenarios. The proposed RL approach facilitates safer and more efficient trajectory planning by enabling the system to anticipate potential rule violations and adjust maneuvers proactively, reducing the likelihood of entering states that lead to unavoidable violations. For rules $R\_{I6}$ and $R\_{G1}$, which involve constraints independent of direct vehicle-to-vehicle interactions or focus on maintaining safe distances, the method exhibits strong performance, as evidenced by the smoother transitions in the value network outputs and the expanded safety regions in the heatmaps.

However, the model shows reduced robustness for rules involving complex vehicle interactions, such as $R\_{I2}$, which requires maintaining appropriate speeds relative to surrounding vehicles. This limitation is reflected in the volatility observed in the explained variance metric and the inconsistent value predictions in the heatmap analysis. The challenge arises from the intricate dynamics of vehicle interactions, which demand a more detailed understanding of relative positions, speeds, and potential maneuvers. Addressing this issue may require enhancements such as increasing the volume of training data, extending the training duration, and incorporating additional features, such as relative velocity or lane-changing probabilities, to better capture the interaction dynamics. Furthermore, adopting a more advanced scenario extraction model, such as a Heterogeneous Graph Neural Network, could enhance the model's ability to represent and manage complex traffic interactions by accounting for diverse vehicles and their interconnected behaviors.

\section{Conclusion}

In this paper, we have introduced an approach to autonomous vehicle path planning that integrates a motion planner with a DRL framework to enhance traffic rule compliance and safety. By replacing the traditional actor network in an AC setup with the cost-based motion planner and leveraging a GNN for state representation, our method effectively combines the interpretability and reliability of motion planning with the predictive power of R. This hybrid framework utilizes the critic’s state-value predictions to guide trajectory selection, enabling the anticipation and prevention of traffic rule violations beyond the immediate planning horizon.

Our approach was evaluated using the highD dataset, focusing on three carefully selected German interstate traffic rules: $R\_G1$, $R\_I2$, and $R\_I6$. Experimental results demonstrate that the proposed method outperforms a non-learning baseline planner, particularly for $R\_I6$ and $R\_G1$, where it achieves smoother value transitions and broader safety margins, as evidenced by heatmap comparisons and metrics such as explained variance and episode reward mean. For $R\_I6$, the model proactively adjusts trajectories to comply with no-overtaking zones, while for $R\_G1$, it anticipates braking events to maintain safe distances. However, performance for $R\_I2$ reveals limitations in handling complex vehicle interactions, exhibiting volatility in predictions that suggests a need for further refinement.

In conclusion, our proposed framework represents a significant step forward in uniting learning-based prediction with explainable planning, offering a scalable and practical solution for autonomous driving. Future work will focus on improving the model’s handling of multi-agent dynamics and exploring its applicability to diverse driving environments beyond German highways.

\bibliographystyle{IEEEtran}
\bibliography{paper}

\end{document}